\documentclass[11pt,letterpaper]{article}
\usepackage{naaclhlt2013}
\usepackage{times}
\usepackage{latexsym}
\usepackage{amsfonts,amssymb,amsmath}
\usepackage{graphicx}
\usepackage[utf8]{inputenc}
\usepackage{url}
\title{How to merge three different methods for information filtering ?}
\author{ \\
	     \\
	     \\
	     \\
	    {\tt  }
	  \And
	 \\
  	 \\
  	 \\
  	 \\
  {\tt  }
\And
	 \\
  	 \\
  	 \\
  	 \\
  {\tt  }
}
\date{}
\begin{document}
\maketitle

\begin{abstract}

Twitter is now a gold marketing tool for entities concerned with online reputation. To automatically monitor online reputation of entities, systems have to deal with ambiguous entity names, polarity detection and topic detection.
We propose three approaches to tackle the first issue: monitoring Twitter in order to find relevant tweets about a given entity. 
Evaluated within the framework of the \textit{RepLab-2013 Filtering task}, each of them has been shown competitive with state-of-the-art approaches. Mainly we investigate on how much merging strategies may impact performances on a filtering task according to the evaluation measure.

\end{abstract}

\section{Introduction}
Online reputation is a key information for public figures and companies in order to react to the public opinions and to anticipate them. Indeed, knowing what make their reputation good or bad allows them to make informed decision. For instance a company may make additional efforts on its call centers if it notices that its consumers are unsatisfied.

Monitoring online reputation of entities requires to be able to retrieve all opinions or reviews about them. Automatic approaches have then to deal with the noise generated by the recall-oriented retrieving techniques used. This noise is mainly the result of entity names ambiguity (e.g. \textit{jaguar} which may refer to an animal or a car manufacturer). A classification step is required to filter out sources which do not actually mention the monitored entity. Topic detection is necessary to identify which matter is discussed in the source and finally the polarity of it has to be estimated (is the opinion positive, neutral or negative?). 
Each of these three issues is an open problem. 
Moreover, systems have to be able to process large amounts of incoming new documents in a short time to provide fresh feedbacks. Sources commonly used are news web sites, blogs, forums or more recently social networks such as Twitter.

We propose three approaches to filter tweets on whether or not they refer to a given entity. These approaches rely on tweets content and meta-data associated to them (timestamp, user names, \ldots) as well as on the information contained about the entity in a knowledge base.
We also investigate if combining systems outputs with merging algorithms can improve the overall performances and if different strategies may be applied to promote a measure or another.

Each proposition is evaluated within the framework provided by the RepLab 2013 evaluation campaign and they all show competitive performances. 


The remainder of this paper is organized as follows. Section~2 presents related works. Section~3 describes the proposed systems. Section~4 gives details about merging algorithms we used. Experiments are described in Section~5. In Section~6 we discuss the results before concluding in section~7.

\section{Related Work}

A decade ago, a TREC task called "Filtering"~\cite{overviewFiltering2002} had the following definition: finding documents relevant to a query in a stream of data. Effective approaches were inspired by information retrieval techniques to score documents (Okapi~\cite{okapiMicrosoft}, Rocchio~\cite{rocchio-boost}, ...). 

In 2012, a new TREC task called "Knowledge Base Acceleration" (\textit{KBA})~\cite{overviewKBA2012} started with a more entity centric definition: filtering a time-ordered corpus for documents that are related to a set of entities from Wikipedia. The best performing approach used one classifier (SVM) by entity tracked with features representing whether or not a term is in a document, regardless of its frequency~\cite{best}. Training data have however to be provided for each new entity "followed". Another successful approach capture intrinsic characteristics of related documents by relying on document centric features, entity profile related features and time features~\cite{Bon:13}. 

Recently information filtering on Twitter emerged. \cite{tweet-eq} for instance followed the evolution of big and short terms events, like natural disasters, in real-time. RepLab 2012 Filtering task~\cite{replab:12} follows the KBA 2012 definition but focus on Twitter as the source of incoming data (instead of news, blogs and forum posts). The submitted approaches rely on various sources of evidence like named entity recognition~\cite{Daedalus:12}, matches of terms between tweets and Wikipedia~\cite{cirgdisco:12} or the importance of features specific to Twitter such as the presence of a user name in a tweet~\cite{ilps:12} or the number of hashtags~\cite{yahoo:12}.

Merging metrics or methods used in natural language processing (\textsc{nlp}) and information retrieval can be seen, as shown in~\cite{Lamontagne2006}, as a multi-criteria optimization problem: in particular, the \textsc{Electre} methods \cite{Fig:05}, which turned out to be efficient applied to industrial domains~\cite{Gou:12}, have been transposed to an \textsc{nlp} context~\cite{Car:12} opted for a voting method to combine their runs with ~\cite{yahoo:12}.

\section{Methods}
\subsection{Cosine distance (TF-IDF-Gini)}
\label{cosine}

The first approach consists in a supervised classification based on a cosine similarity. Vectors used to compute similarities are built using the Term Frequency-Inverse Document Frequency (TF-IDF)~\cite{Sal:88} and the Gini purity criterion~\cite{Tor:12}.

Tweets are cleansed by removing hypertext links and punctuation marks, hashtags and @ before a user name. We have removed a set of tool-words and some entities ID. Terms are lower-cased. We generate a list of n-grams by using the Gini purity criterion. 

We create terms (words or n-grams) models for both classes (related and unrelated tweets) and term frequencies are computed with the TF-IDF and Gini criterion. 
These models take into account the following meta-data: user id, entity id and language integrated as terms in the bag-of-terms of tweets.

A cosine similarity measures the distance between the bag-of-terms of a tweet and the whole bag built for each class and ranks tweets according to this measure. 

\subsection{KNN with discriminant features}

The system tries to match each tweet in the test set with the $\textit{K}$ most similar tweets in the training set. 
Tweet similarity is computed using Jaccard measure on the bag-of-words discriminant representation of the tweets.
As in section~\ref{cosine}, each tweet is represented as a vector whose components are weighted according to TF-IDF and the Gini purity criterion.
The process also takes into account tokens created from the meta-data (author, entity-id).
The stoplist of section~\ref{cosine} has been used.

\subsection{Adaptation KBA'12 system}

For the KBA filtering task, a state-of-the-art approach consist in capturing intrinsic characteristics of highly relevant documents by mean of three types of features: document centric features, entity's profile features, and time features~\cite{Bon:13}.
Features are computed for each candidate document and, using a Random Forest classifier, used to determine if the document is related or not to a given entity. 

Unlike previous approaches it doesn't require a new set of examples for each new entity.
We want to measure the robustness of this approach by using it on another type of documents (i.e. tweets). No adjustments are made on it but tweets are however preprocessed: stop-words are deleted as well as \textit{@} before user names and hashtags are split. 
The classifier is trained on all related and unrelated examples for each type of entities (automotive, universities, banking and music/artists).

\section{Merging algorithms}

To improve the performances we use three ways of combining our systems outputs.

\subsection{Linear combination of outputs score}
$N$ systems are available. For each tweet $T$ of the test set, one system $j$ associates each label $L_k$ with a confidence score $s_j(T,L_k) \ (j = 1, ..., N)$.  The output entity label L is chosen according to the following rule~:

\begin{equation}
  L = {\operatorname{arg\,max}}_k \left ( \sum_{j=1}^N s_j(T, L_k) \right )
\end{equation}

\subsection{\textsc{Electre}~I method}

The goal of this method~\cite{Roy:91} is to choose the best label from the entire set of labels ranked according to the different systems.

A relation $\mathcal{S} \subset \mathbb{L} \times \mathbb{L}$, denoted ``over-ranking'', is defined on the label set $\mathbb{L}$: a label
$l$ over-ranks another label $l'$ if $l$ dominates $l'$ on an ``important'' number of systems and if $l'$ does not dominate ``too much'' $l$ on the remaining systems.

More precisely, for each pair of labels $(l, l')$, a concordance index $c(l, l')$ is computed, corresponding to the proportion of systems
where $l$ dominates $l'$. $l$ over-ranks $l'$ if $c(l, l')$ exceeds a concordance threshold, generally fixed around $2/3$ and if $l$ is not
dominated by $l'$ on the remaining systems above a \textit{veto} threshold, which has been fixed here to $v = 0.5$.

The set of the best labels, possibly empty and denoted as the kernel
of the relation $\mathcal{S}$, consists in the labels which are not
overanked by others. If there is no, or more than one, label in the
kernel, this method is discarded and the merging algorithm described
in the previous subsection, based on a linear combination of the
scores, is applied.

\subsection{\textsc{Promethee} mono-criterion method}

This method relies on a concordance matrix: for each pair of labels $(l_i, l_j)$, the matrix coefficient $c_{ij}$ corresponds to the
concordance index $c(l_i, l_j)$ introduced in the previous subsection.

For each label $l_i$, two sums are computed: $s_l(l_i) = \sum_j c_{ij}$ and $s_c(l_i) = \sum_j c_{ji}$. $s_l(l_i)$ measures the
tendency of $l_i$ to dominate the other labels, and $s_l(l_i)$ the tendency of $l_i$ to be dominated.

The final score of the label $l_i$ is the difference $s_l(l_i) -
s_c(l_i)$ and the dominant label is the one whose score is maximal.

\section{Experiments}

\subsection{Replab 2013 Framework}

The corpus is a bilingual (English and Spanish) collection of tweets containing the name of one of the 61 entities selected in four domains: automotive, banking, universities and music/artists.
Tweets have been collected by querying the Twitter search engine\footnote{http://twitter.com/search}.
The dataset covers a period going from the 1$^{\text{st}}$ of June 2012 to the 31$^{\text{st}}$ of December 2012.
42,700 tweets have been provided for training purpose and 100,000 tweets for the evaluation. The training set is composed of the 700 first tweets retrieved for each entity. 
For each entity, at least 2,200 tweets have been collected. 

Tweets, however, are not homogeneously distributed across the entities. 

Systems are evaluated according to the following measures: Accuracy, Reliability and Sensitivity ~\cite{Ami:13}. 
Reliability is defined as precision of binary relationships predicted by the system with respect to those that derive from the gold standard; and Sensitivity is similarly defined as recall of relationships.
A F-measure is then used to combine both scores.

These measures are well adapted to the task but are really severe on unbalanced datasets. 

\subsection{Results}
Table~\ref{results} shows  results of our approaches against the official RepLab 2013 baseline and the median system among participants.
\begin{table*}[!ht]
 \begin{center}
   \tabcolsep = 2\tabcolsep 
   \begin{tabular}{lccccc}
   \hline
		Approach & Accuracy & Reliability & Sensitivity & F-Measure \\
   \hline
   MPMS & .899 & .668 & \textbf{.367} & \textbf{.400} \\
   OTB & .902 & .651 & \textbf{.367} & .386 \\
   Naive LC & \textbf{.904} & \textbf{.691} & .364 & .385 \\
   Naive Elec & .903 & .671 & .363 & .383 \\
   k-NN & .890 & .658 & .357 & .381 \\
   KBA & .878 & .619 & .331 & .341\\
   R-LC & .895 & .680 & .290 & .313\\   
   \textit{Baseline} & \textit{.876} & \textit{.461} & \textit{.325} & \textit{.312} \\
   R-Elec & .892 & .680 & .281 & .302\\
   Cosine & .834 & .423 & .331 & .272\\
   \textit{RepLab Median} & \textit{ .826} & \textit{.489} & \textit{.286 }& .\textit{.265}\\
   \hline
   \end{tabular}
\caption{Results on the Filtering Task ordered according to the F-Measure.} 
 \label{results}
 \end{center}
\end{table*} 

The baseline\footnote{http://www.limosine-project.eu/events/replab2013} is a supervised system that matches each tweet in the test set with the most similar tweet in the training set, and assumes that the annotations in the tweet from the training set are also valid for the tweet in the test set. Tweet similarity is computed using Jaccard distance and a straightforward bag-of-words representation of the tweets.

The method described in section 3.2 can be considered as an improved version of the baseline.

Two systems (KNN and KBA with a F-measure scores of respectively $.381$ and $.341$) have reached greater performances than the baseline on every measures.
The confidence interval (.002 and .005 respectively for accuracy and F-Measure) computed following \textit{Polling Method} ~\cite{Voo:98} shows that the difference between the systems is significant. 

Merging strategies R-Elec (for \textsc{Electre}) and R-LC (for Linear Combination) did not produce good selection rules since their performances remain lower than the best system taken alone.
A natural merging strategy consisting in merging only the best systems on a development set gives better results (Naive LC and Naive Elec). 

Moreover, a multi pass strategy (MPMS) merging systems in pair before considering merging all pairs improves Sensitivity and thus the F-Measure (.400) despite of a loss in term of accuracy and reliability. 
 Finally, merging only the best (OTB) runs on each measure gives quite similar improvements.

These results show that using merging strategies to combine different systems lead to improvements, whatever the metric chosen. The key observation is that it is possible to pick a merging strategy according to the metrics we choose to focus on. A quite naive merging strategy (Naive) seems to result in a better precision (improvements in both Accuracy and Reliability). On the contrary, adopting a multi pass strategy (MPMS) allow to give a highest priority to recall in both classes (i.e. Sensitivity). Finally, if a compromise is preferred, we saw that promoting systems that did well on each measure (OTB) is a good option.




\section{Conclusion}
In this paper we presented some of the interesting features of the systems that we evaluated within the framework provided by RepLab 2013 as well as their performances.
We proposed several combinations of them using different merging strategies in order to take benefit from the diversity of information offered by our systems.
We also showed that these merging strategies have to be applied depending on the evaluation measures to offer in one hand the best results according to a specific measure or in the other hand to obtain a trade-off.
Since a merging strategy cannot get the best score according to each metrics, we can accept a loss according to one metric if it has a real impact on the task official measures. 

A more advanced view would be to apply a specific merger entity by entity, especially for unbalanced entities.

\end{document}